\documentclass[12pt,a4paper,twoside]{article}

\usepackage{times}
\usepackage[latin1]{inputenc}
\usepackage[T1]{fontenc}

\usepackage{graphicx}

\usepackage{taln2009Poster} 
\usepackage{wrapfig}
\usepackage[frenchb]{babel}
\newcommand{\fup}[1]{\up{#1}}

\renewcommand\sectionmark[1]{}
\renewcommand\subsectionmark[1]{}
\renewcommand\refname{}

\newcommand{\french}[1]{{\emph{``#1''}}}

\newcommand{\node}[1]{\texttt{#1}}

\begin{document}

\title{Analyse en dépendances à l'aide des grammaires d'interaction}

\author{Jonathan Marchand\fup{1} 
  \quad Bruno Guillaume\fup{2}
  \quad Guy Perrier\fup{1}\\
  (1) LORIA~/~Université Nancy~2\\ 
  (2) LORIA~/~INRIA Nancy Grand-Est \\ 
  prénom.nom@loria.fr\\ 
} 

\date{}

\maketitle

\resume{
Cet article propose une méthode pour extraire une analyse en
dépendances d'un énoncé à partir de son analyse en constituants avec
les grammaires d'interaction. Les grammaires d'interaction sont un
formalisme grammatical qui exprime l'interaction entre les mots à
l'aide d'un système de polarités. Le mécanisme de composition
syntaxique est régi par la saturation des polarités. Les interactions
s'effectuent entre les constituants, mais les grammaires étant
lexicalisées, ces interactions peuvent se traduire sur les mots. La
saturation des polarités lors de l'analyse syntaxique d'un énoncé
permet d'extraire des relations de dépendances entre les mots, chaque
dépendance étant réalisée par une saturation. Les structures de
dépendances ainsi obtenues peuvent être vues comme un raffinement de
l'analyse habituellement effectuée sous forme d'arbre de
dépendance. Plus généralement, ce travail apporte un éclairage nouveau
sur les liens entre analyse en constituants et analyse en dépendances.
}

\abstract{This article proposes a method to extract dependency
  structures from phrase-structure level parsing with Interaction
  Grammars. Interaction Grammars are a formalism which expresses
  interactions among words using a polarity system. Syntactical
  composition is led by the saturation of polarities. Interactions
  take place between constituents, but as grammars are lexicalized,
  these interactions can be translated at the level of words. Dependency
  relations are extracted from the parsing process: every dependency
  is the consequence of a polarity saturation.
  The dependency relations we obtain can be seen as a refinement of
  the usual dependency tree. Generally speaking, this work sheds new
  light on links between phrase structure and dependency parsing. 
 }

\motsClefs{Analyse syntaxique, grammaires de dépendances, grammaires
d'interaction, polarité}
{Syntactic analysis, dependency grammars, interaction grammars, polarity}

\newpage
\section{Introduction}
Les grammaires de constituants et les grammaires de dépendances sont
souvent présentées comme orthogonales~: les premières organisent les
groupes de mots en syntagmes alors que les secondes mettent la
dépendance entre mots au centre de la structure syntaxique. Avec les
grammaires de constituants lexicalisées, telles que les grammaires
d'arbres adjoints (TAG) et les grammaires catégorielles (CG), où
chaque élément de la grammaire est associé à un mot, la composition
syntaxique lors de l'analyse met en évidence des liens entre les
mots. Ces liens présentent des similitudes avec les relations de
dépendances et ont fait l'objet de différentes études.

A.~Dikovsky et L.~Modina ont étudié du point de vue formel le passage
d'une analyse en constituants à une analyse en dépendances et vice
versa~\cite{dikovsky00}. O.~Rambow et A.~Joshi ont expliqué comment
retrouver une analyse en dépendances à partir d'une analyse dans les
TAG où les substitutions et les adjonctions sont vues comme des
relations de dépendances entre les mots~\cite{rambow94}. Enfin,
l'article~\cite{Clark02} propose une méthode similaire pour les grammaires
catégorielles combinatoires où l'application des règles combinatoires
donne lieu à des relations de dépendances entre les mots.

Les grammaires d'interaction (IG) sont des grammaires de constituants
lexicalisées qui étendent par un système de polarités plus riche le
système besoins/ressources employé dans les grammaires catégorielles.
Dans cet article, nous généralisons, les résultats~\cite{Clark02}
cités plus haut pour les CG au cas des IG, en révisant la méthode~: en
effet, cette dernière impose d'étendre les entrées lexicales avec des
marqueurs pour aider à la construction des dépendances lors de
l'analyse, et produit trop de dépendances, alors que notre méthode
s'appuie plus simplement sur le lien entre polarités et dépendances.

Dans la section~\ref{sec-gi}, les IG sont présentées et illustrées par
un exemple. La section~\ref{sec-analyse} décrit la méthode
d'extraction des dépendances à partir d'une analyse avec une IG. Finalement,
dans la section~\ref{sec-structures}, nous étudions les structures de
dépendances obtenues et nous mettons en perspective avec d'autres travaux.

\section{Les grammaires d'interaction}
\label{sec-gi}
Les grammaires d'interaction \cite{Per2003} sont un formalisme
grammatical s'appuyant sur la notion de \textit{description
  d'arbres}. Cette notion a été introduite par M.~Marcus, D.~Hindle et
M.~Fleck en 1983 \cite{Marcus83}, et K.~Vijay-Shanker l'a utilisée
pour représenter de façon monotone l'opération d'adjonction des
TAG \cite{Vij92}.

Une description d'arbres est définie par un ensemble de n\oe uds et de
relations d'ascendance, de parenté et de précédence entre ces n\oe
uds. Les n\oe uds représentent des syntagmes (éventuellement vides) et
les relations expriment les dépendances entre ces syntagmes. Les
propriétés morpho-syntaxiques de ces syntagmes sont décrites par des
structures de traits.

Cette approche flexible est bien adaptée à l'ambiguïté des langues
naturelles. Cependant, l'analyse syntaxique fondée sur des
descriptions d'arbres est très coûteuse \cite{KNR01}. En effet,
dans cette approche, l'analyse syntaxique consiste à chercher des
modèles de descriptions d'arbres sous forme d'arbres syntaxiques
complètement spécifiés, ce qui est un problème NP-complet.

Dans les formalismes opérationnels fondés sur les descriptions
d'arbres (comme les D-tree substitution grammars \cite{ROD01} ou les
TT-MCTAG \cite{Kallmeyer03}), cet indéterminisme est limité en
contraignant la syntaxe des descriptions et le mécanisme de
composition syntaxique.

L'originalité des grammaires d'interaction est de proposer un
mécanisme de composition syntaxique très souple qui consiste à
superposer les descriptions d'arbres mais qui est guidé par une
contrainte de saturation de \textit{polarités}.  Cette contrainte fait
référence à l'idée de valence de Tesnière \cite{Tesniere:1934} et est
essentielle dans les CG~: chaque mot est équipé d'une valence
exprimant ses possibilités d'interaction avec les autres mots. La
composition syntaxique est contrôlée pas une dualité
besoins/ressources~: certaines ressources munies de polarités
négatives sont attendues alors que d'autres, munies de polarités
positives, sont disponibles. Dans les IG, cette idée de valence est
reprise et généralisée.

\subsection{Système de polarités}

Contrairement aux CG, les IG attachent les polarités aux traits qui
décorent les n\oe uds. Mais nous nous en tiendrons ici à une version
simplifiée des IG où les polarités sont accrochées aux n\oe uds. Une
autre différence avec les CG est que le système de polarités est plus
riche. En effet, les IG proposent deux types d'interaction à base de
polarités:
\begin{itemize}
\item \textbf{les interactions linéaires~:} chaque n\oe ud portant une
  polarité positive (notée $+$) doit fusionner avec exactement un n\oe
  ud portant une polarité négative (notée $-$) et réciproquement.
\item \textbf{les interactions non-linéaires~:} chaque n\oe ud portant
  une polarité virtuelle (notée $\sim$) doit fusionner exactement,
  soit avec un n\oe ud positif et un n\oe ud négatif, soit avec un
  n\oe ud portant la polarité saturée (notée $=$). En revanche, un
  nombre quelconque de noeuds virtuels peuvent fusionner avec le même
  couple positif/négatif ou avec le même n\oe ud saturé.

\end{itemize}

\begin{wraptable}{r}{10em}
\small{    \begin{tabular}{|c | c c c c |}
      \hline
      & $\sim$ & $-$ & $+$ & $=$  \\
      \hline
      $\sim$ & $\sim$ & $-$ & $+$ & $=$ \\
      $-$ & $-$ & & $=$ & \\
      $+$ & $+$ & $=$ & &  \\
      $=$ & $=$ & & &  \\
      \hline
    \end{tabular}}
\end{wraptable}
Lors de la fusion de deux n\oe uds, le n\oe ud résultant porte la
polarité issue de la composition des polarités des n\oe uds
initiaux. La composition d'une polarité positive et d'une polarité
négative donne une polarité saturée alors que la polarité virtuelle
est l'élément neutre de cette opération.  Toute autre composition
provoque l'échec de la fusion
(cf. tableau ci-contre). L'opération de composition est
associative et commutative, l'ordre de fusion des n\oe uds n'a donc
pas d'importance dans le processus de composition syntaxique.

\subsection{Analyse avec les IG}

La structure syntaxique élémentaire manipulée dans les IG est appelée
{\bf description d'arbre polarisée} (DAP). Une IG particulière est
définie par un ensemble de DAP~; chaque DAP est associée à un
mot\footnote{Dans nos grammaires, il n'y a pas de co-ancre et donc un
  seul mot par DAP (l'usage de co-ancre peut-être simulé en utilisant
  des polarités).}. La grammaire est ainsi un lexique où un mot peut
avoir plusieurs entrées. Pour analyser une phrase, il faut choisir
pour chaque mot l'une des DAP associée à ce mot. Un tel choix est
appelée une {\bf sélection lexicale}. L'analyse consiste ensuite à
composer ces DAP pour obtenir un arbre d'analyse.

L'opération atomique de composition syntaxique consiste à superposer
deux n\oe uds pour saturer leurs polarités. On itère l'opération de
saturation de n\oe uds pour construire progressivement l'analyse de la
phrase sous forme d'un arbre.

\begin{figure}[htbp]
\begin{center}
\includegraphics[scale=0.32]{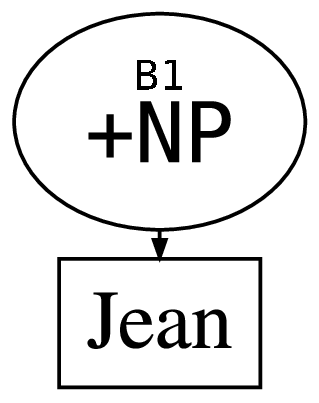}
\includegraphics[scale=0.32]{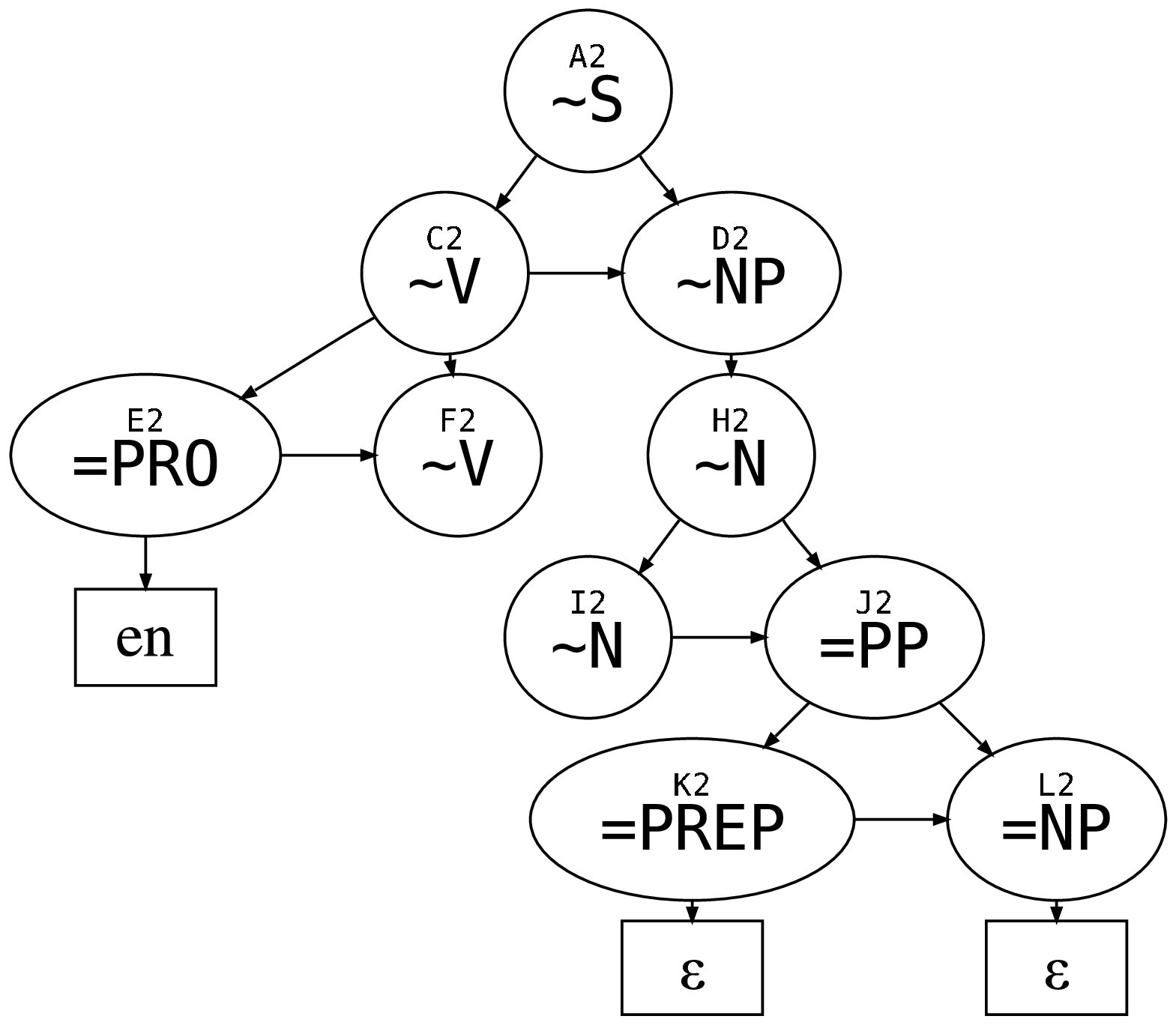}
\includegraphics[scale=0.32]{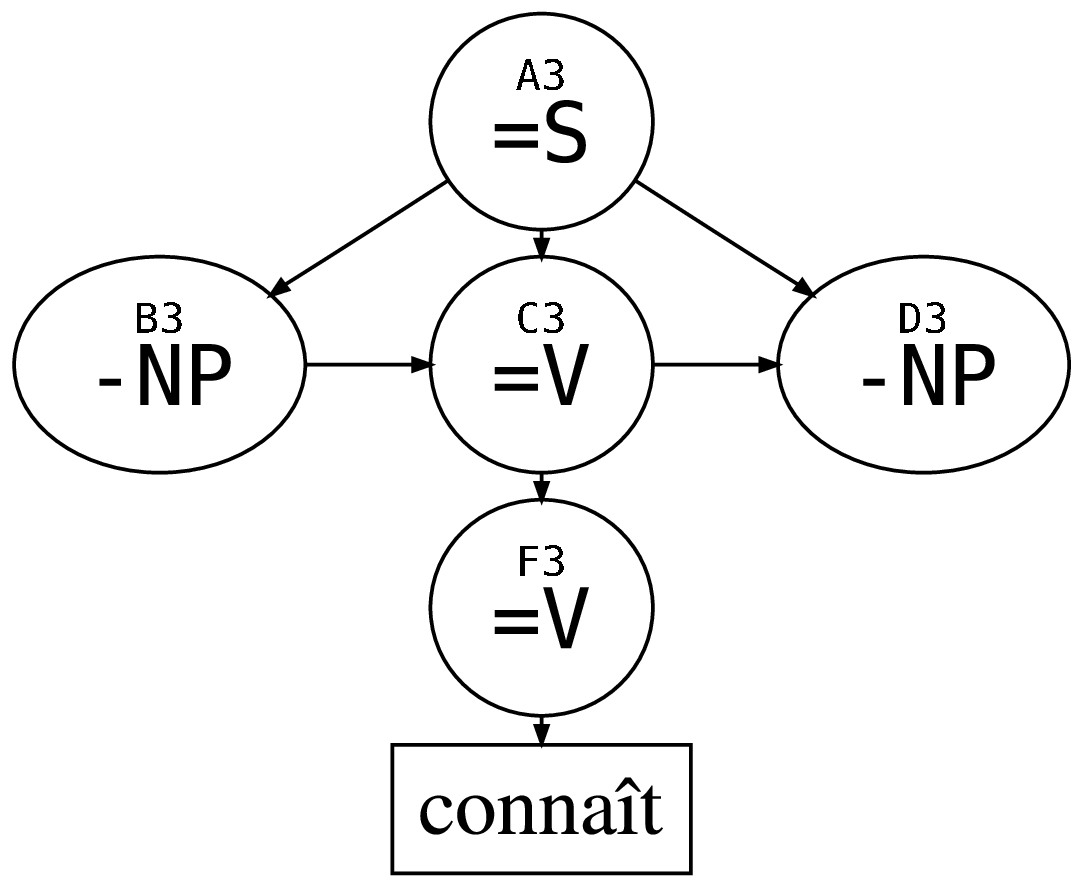}
\includegraphics[scale=0.32]{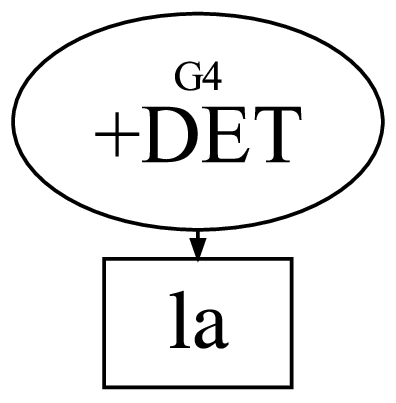}
\includegraphics[scale=0.32]{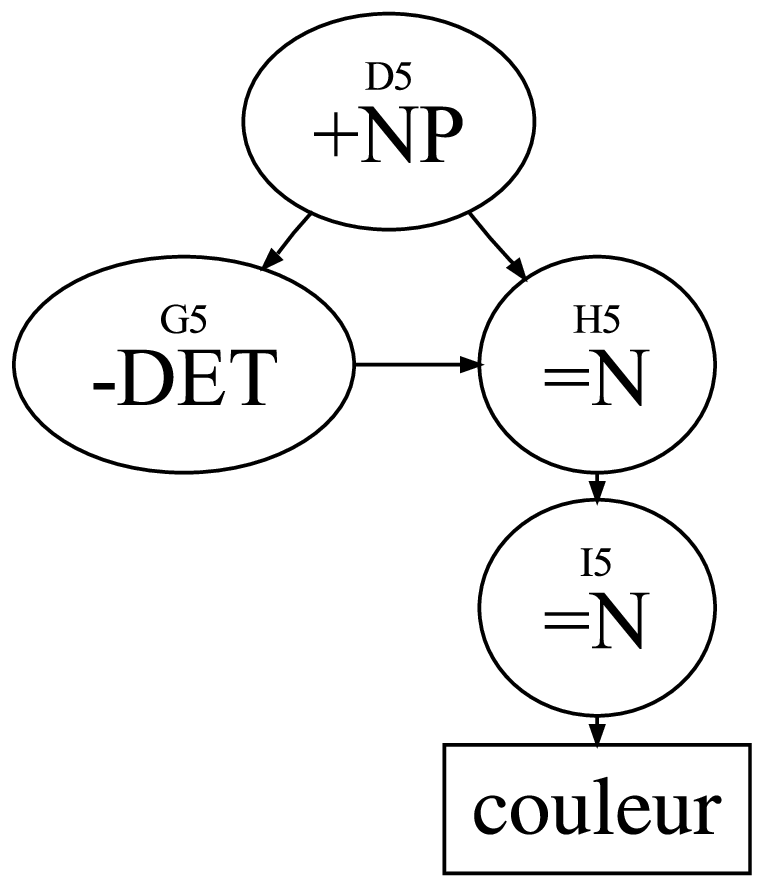}
\end{center}
\caption{DAP associées à la phrase \french{Jean en connaît la
  couleur}\label{selection}}
\end{figure}

Les DAP d'une sélection lexicale pour la phrase \french{Jean en connaît la
couleur} sont représentées sur la figure~\ref{selection}.

La DAP représentant le mot \french{en} décrit l'intuition
linguistique suivante~: le pronom \french{en} est utilisé comme complément
du nom \french{couleur} mais il vient s'adjoindre devant le verbe
\french{connaît} qui admet \french{la couleur} comme objet direct. La DAP montre
à droite du noyau verbal \node{C2} un syntagme objet \node{D2}
attendu comportant un complément du nom \node{J2} qui est déjà
complètement saturé mais sans réalisation phonologique. Le n\oe ud
\node{I2} renseigne la positionnement du nom dans le syntagme
\node{H2}. Au niveau du noyau verbal \node{C2}, le pronom \french{en}
\node{E2} est positionné à gauche du verbe \node{F2}.

\begin{figure}[htbp]
\begin{center}
\includegraphics[scale=0.32]{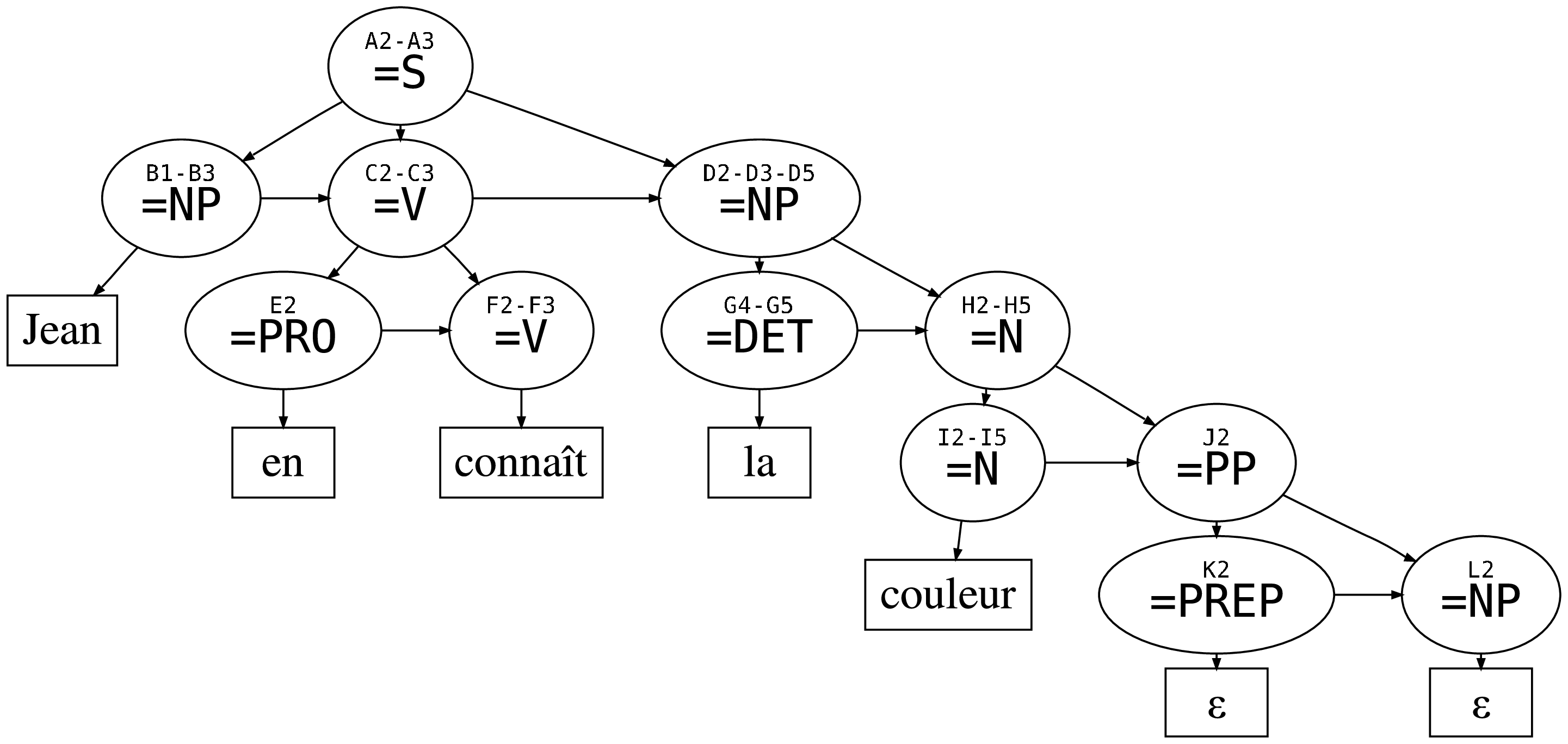}
\end{center}
\caption{Arbre d'analyse de \french{Jean en connaît la couleur}\label{modele}}
\end{figure}

À partir de l'ensemble des DAP de la figure~\ref{selection}, on peut
construire l'arbre syntaxique saturé représenté sur la
figure~\ref{modele}. Sur chaque n\oe ud de l'arbre est indiqué
l'ensemble des n\oe uds des DAP qui ont été superposés. Par exemple le
n\oe ud \node{A2-A3} représente la superposition du n\oe ud \node{A2} de la
DAP de \french{en} et du n\oe ud \node{A3} de la DAP de
\french{connaît}.
 

\subsection{Grammaire du français}
Pour valider le formalisme, nous avons développé une grammaire du
français~\cite{perrier07}. Cette grammaire a été évaluée sur la
TSNLP~\cite{Lehmann96tsnlp} qui contient 1690 énoncés grammaticaux et
1935 énoncés agrammaticaux. Ce jeu de tests ne couvre pas toute la
langue française, il y a peu de phrases complexes mais il insiste sur
certains phénomènes comme la coordination ou la position des
compléments adverbiaux. Cependant, notre grammaire couvre d'autres
phénomènes dont la TSNLP ne tient pas compte, comme par exemple~: la
voix passive, la sous-catégorisation des noms et des adjectifs
prédicatifs, le contrôle du sujet des compléments infinitifs, les
propositions relatives et interrogatives. 88\% des 1690
phrases grammaticales sont modélisées et 85\% des 1935 phrases
agrammaticales sont rejetées. Les 15\% d'énoncés agrammaticaux sont
acceptés car la grammaire ne modélise pas les règles phonologiques et
la sémantique. Les raisons pour lesquelles 12\% des énoncés
grammaticaux ne sont pas analysés sont diverses (phrases
retranscrites de l'oral, expressions figées, causatifs, superlatifs).

\section{Analyse en dépendances}
\label{sec-analyse}
L'analyse syntaxique est obtenue par superposition des DAP d'une
sélection lexicale. La superposition est guidée par la fusion des n\oe
uds portant des polarités se saturant. Au niveau d'une DAP, ces
polarités représentent les besoins/ressources du mot dans un énoncé.
La saturation de ces dernières peut alors se voir comme la résolution
d'une dépendance entre ces mots. On peut alors retrouver les relations
de dépendances d'une phrase à partir de l'ensemble des DAP associées
aux mots de la phrase et de son analyse.

\subsection{Dépendances linéaires}
Nous nous intéressons dans un premier temps au cas simple de la
saturation linéaire de deux polarités $+$ et $-$. Par exemple, dans la
phrase \french{Jean en connaît la couleur} (Figure~\ref{modele}), la
DAP représentant le déterminant \french{la} possède un n\oe
ud~\node{G4} portant une polarité positive et étiqueté par la
catégorie syntaxique {\tt DET}.  La DAP du nom \french{couleur}
comporte quant à elle un n\oe ud~\node{G5} portant une polarité
négative qui est aussi étiqueté par {\tt DET}.

Lors de l'analyse, ces deux n\oe uds fusionnent pour obtenir un n\oe
ud saturé, la DAP résultante représentant une analyse partielle de
\french{la couleur}. La saturation de ces deux n\oe uds peut être vue
comme la réalisation d'une relation de dépendances entre les deux mots
correspondants.

L'opération de saturation représente la satisfaction d'une contrainte
de besoins/ressources. Un élément qui se présente comme nécessitant une
ressource est considéré comme le gouverneur de la relation de dépendances,
tandis qu'un élément qui se présente comme fournissant une ressource
se retrouve comme le dépendant de cette relation. Ainsi,
en cas d'interaction linéaire, le mot dont la DAP contient le n\oe ud
négatif est le \textit{gouverneur} et le mot dont la DAP contient le
n\oe ud positif est le \textit{dépendant} de la relation de dépendances.

Les relations de dépendances engendrées par la saturation des n\oe uds
portant des polarités opposées seront appelées {\bf dépendances
  linéaires}. Dans la grammaire actuelle du français, elles
représentent les relations tête-complément et tête-spécifieur.
\begin{figure}[htbp]
\begin{center}
\includegraphics[scale=1.0]{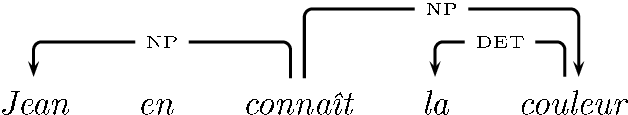}
\caption{Relations de dépendances linéaires dans la phrase \french{Jean en
connaît la couleur} \label{enconnaitlinéaire}}
\end{center}
\end{figure}

Dans l'exemple \french{Jean en connaît la couleur}, les dépendances
linéaires obtenues sont représentées sur la
figure~\ref{enconnaitlinéaire} (les arcs portent la catégorie qui a
fait l'objet d'une saturation\footnote{en pratique, il est possible
  d'utiliser d'autres étiquettes sur les arcs comme les fonctions
  syntaxiques (elles sont indiquées dans les structures de traits).}).

Cette analyse possède un n\oe ud \french{en} isolé. La DAP du pronom
\french{en} ne porte en effet pas de polarité positive ou négative et
ainsi ne produit pas de relation de dépendances linéaires avec le
reste de la phrase. La saturation des polarités positives et négatives
ne suffit donc pas pour exprimer toutes les relations de dépendances
d'une phrase. Nous allons donc voir comment certaines relations de
dépendances peuvent être produites pas des interactions non-linéaires.

\subsection{Dépendances non-linéaires}


\begin{wrapfigure}{r}{25mm}
\begin{center}
\includegraphics[scale=0.36]{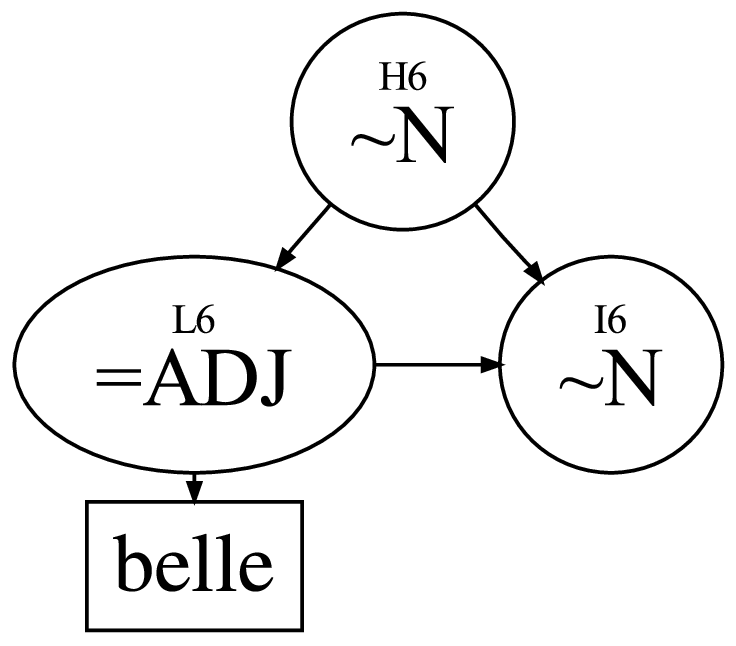}
\end{center}
\end{wrapfigure}
Dans la grammaire actuelle, la DAP pour l'adjectif \french{belle} dans
le groupe nominal \french{la belle couleur} est donnée par la figure ci-contre.
De façon habituelle en dépendances, on
considère qu'il y a une dépendance de \french{belle} vis à vis de
\french{couleur}. Deux n\oe uds sont non-saturés (\node{H6} et
\node{I6}) et ils portent tous les deux une polarités virtuelle,
c'est donc la saturation de l'une de ces deux polarités qui doit
induire la dépendance. Dans ce cas, les deux polarités peuvent être à
l'origine de la dépendance. Cependant, il ne doit être produit qu'une
seule relation de dépendances, il faut choisir alors quelle polarité
engendrera une dépendance lors de sa saturation.

\begin{wrapfigure}{r}{35mm}
\begin{center}
\includegraphics[scale=0.36]{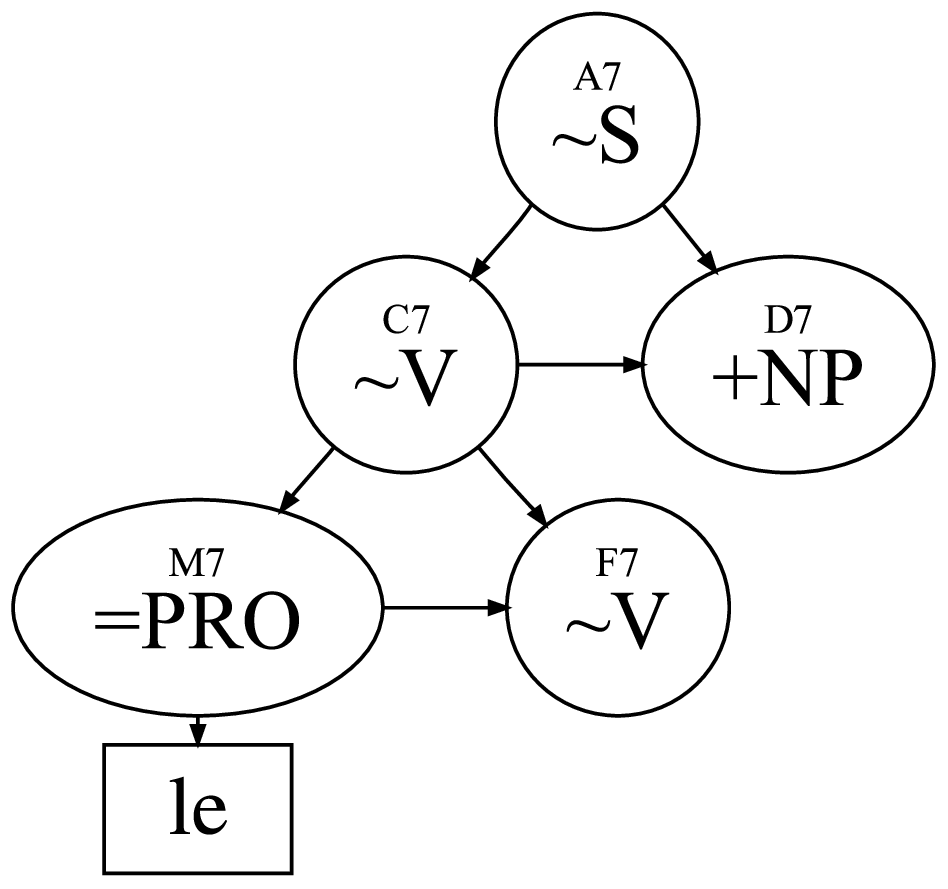}
\end{center}
\end{wrapfigure}
Un autre exemple d'usage des polarités virtuelles dans la grammaire du
français est illustré par l'exemple \french{Jean le connaît}. Dans
cette phrase, le pronom \french{le} (ci-contre) est l'objet
direct du verbe \french{connaît}~: cette relation de dépendances est
produite par la saturation linéaire des polarités du n\oe ud \node{D7}
de la DAP du pronom \french{le} et du n\oe ud \node{D3} de la DAP du
mot \french{connaît}. Mais la DAP du pronom \french{le} possède trois
n\oe uds \node{A7}, \node{C7}, \node{F7} portant une polarité
virtuelle dont la saturation non-linéaire n'apporte aucune information
de dépendances entre les mots \french{le} et \french{connaît}. Ces
polarités contrôlent simplement la place des syntagmes lors de la
superposition des deux DAP et gèrent le fait que le pronom \french{le}
se place avant le verbe alors que la place canonique du groupe nominal
objet dans la phrase est après le verbe. Dans l'arbre d'analyse de
\french{Jean le connaît} les trois n\oe uds virtuels \node{A7},
\node{C7}, \node{F7} sont saturés, respectivement, par les trois n\oe
uds \node{A3}, \node{C3}, \node{F3} de la DAP du mot \french{connaît}.

Les deux derniers exemples montrent bien qu'il y a deux usages
distincts des polarités virtuelles qui ne se comportent pas de la même
manière pour la production de relation de dépendances~:
\begin{itemize}
\item les {\bf polarités virtuelles de dépendances} qui portent une information
  sur les relations de dépendances d'un mot avec son environnement~;
\item les {\bf polarités virtuelles de contexte} qui imposent des contraintes sur
  le contexte syntaxique d'un mot.
\end{itemize}

Il n'est pas possible de distinguer automatiquement, dans une
grammaire donnée, les deux types de polarités virtuelles. C'est donc à
l'auteur de la grammaire de se baser sur des critères linguistiques
pour distinguer ces deux usages. Cependant, dans la pratique,
l'utilisation de métagrammaires (notre grammaire, par exemple, est
construite avec XMG~\cite{xmg}) permet de faire ce travail rapidement,
de façon cohérente sur l'ensemble de la grammaire. Cela concerne
uniquement les mots jouant le rôle de modificateurs (adjectifs
épithètes, adverbes, prépositions introduisant des compléments
adjoints, pronoms relatifs dans leur rôle par rapport à leur
antécédent, pronom clitique "il" utilisé en redoublement du
sujet, etc.). Il s'agit, dans la DAP associé au mot concerné de marquer le
n\oe ud considéré comme n\oe ud privilégié de rattachement au mot qui
est modifié. Le travail a été effectué sur notre grammaire du français à
large couverture en moins d'une heure.

Ainsi, dans la DAP de \french{en} (figure~\ref{selection}), toutes les
polarités virtuelles des n\oe uds \node{A2}, \node{C2},
\node{D2} et \node{F2} sont des polarités virtuelles de contexte
qui gèrent le positionnement de \french{en}. Pour rendre compte du
fait que \french{en} dépend de \french{couleur}, il faut que l'une des
deux polarités virtuelles \node{H2} ou \node{I2} soit une polarité
virtuelle de dépendances. Dans notre cas, nous avons choisi
arbitrairement la polarité \node{H2}.

Dans le cas où une relation de dépendances est produite, le
\textit{dépendant} est le mot dont la DAP porte le n\oe ud virtuel de
dépendances. La polarité virtuelle peut être saturé~:
\begin{itemize}
\item soit par un n\oe ud portant la polarité $=$, dans ce cas ce
  n\oe ud est le gouverneur~;
\item soit par un couple de n\oe ud (un positif et un négatif), dans
  ce cas le gouverneur est le n\oe ud portant la polarité positive.
\end{itemize}

Les relations ainsi engendrées sont appelées {\bf dépendances
  non-linéaires}. En effet, un n\oe ud saturé pouvant se composer avec
zéro ou plusieurs n\oe uds virtuels, plusieurs relations de dépendances
peuvent avoir comme gouverneur le même mot. Ces relations de dépendances
expriment généralement une relation modifieur-modifié.

\begin{figure}[htbp]
\begin{center}
\includegraphics[scale=1.0]{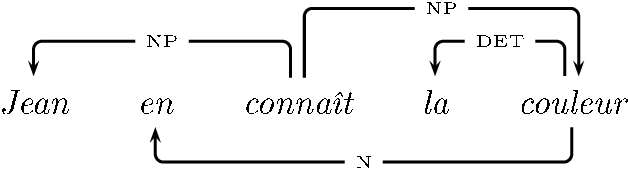}
\caption{Relations de dépendances dans la phrase \french{Jean en
 connaît la couleur} \label{enconnait}}
\end{center}
\end{figure}

Dans les structures de dépendances (figures~\ref{enconnait} ci-dessous
et~\ref{la_fille_que_} plus loin) les dépendances linéaires sont
représentées au-dessus de la phrase et les non-linéaires au-dessous.

La procédure d'extraction des dépendances se résume ainsi:

\begin{itemize}
\item La saturation de polarités opposées engendre une relation de
 dépendances linéaire entre les mots~; la relation va de la polarité
 négative vers la polarité positive.
\item La saturation d'une polarité virtuelle de dépendances avec une
  polarité positive ou saturée engendre une relation de dépendances non
  linéaire entre les mots~; la relation va de la polarité saturée ou
  positive vers la polarité virtuelle de dépendances.
\end{itemize}

Cette procédure permet d'obtenir l'analyse en dépendances de
\french{Jean en connaît la couleur} représentée figure~\ref{enconnait}.

\vspace{0.8cm}

Pour garantir que les structures de dépendances sont connexes (chaque
mot est en relation avec au moins un autre mot de l'énoncé), il suffit
d'imposer que chaque DAP de la grammaire contienne au moins un n\oe ud
positif, un n\oe ud négatif ou un n\oe ud portant une polarité
virtuelle de dépendances. C'est la cas de la grammaire actuellement
implantée pour le français.

\section{Structures de dépendances obtenues}
\label{sec-structures}

Le choix du type de structures pour représenter les dépendances d'une
phrase est une question épineuse qui divisent les linguistes. L'idée
de départ des grammaires de dépendances est de considérer que chaque
mot de la phrase (sauf le verbe principal) est {\em gouverné} par
exactement un autre mot de la même phrase. Cette hypothèse conduit à
considérer que les bonnes structures de dépendances sont les arbres,
nous allons voir ce qu'il en est avec notre méthode qui permet
d'observer, sans a priori, les structures obtenues.

\subsection{Graphes orientés}
De façon générale, la structure en dépendances que l'on obtient est un
graphe orienté~; de plus, avec la restriction imposée sur les DAP de
la grammaire, on sait que ce graphe est connexe.

\begin{figure}[htbp]
\begin{center}
\includegraphics[scale=1.0]{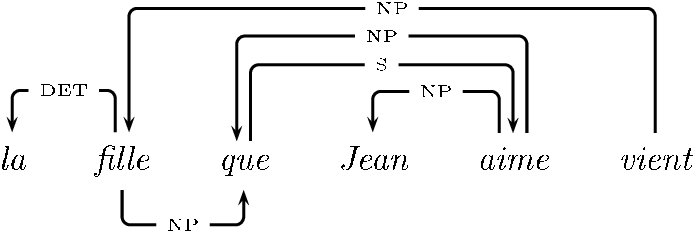}
\caption{Relations de dépendances dans la phrase \french{la
fille que Jean aime vient} \label{la_fille_que_}}
\end{center}
\end{figure}

L'analyse en dépendances pour la phrase \french{Jean en connaît la
  couleur} donnée par la figure~\ref{enconnait} est un arbre~; en
effet, tous les mots, sauf \french{connaît}, ont un et un seul
gouverneur.  Il existe cependant des exemples pour lesquels la
structure de dépendances n'est pas un arbre. L'application de notre
méthode à la phrase \french{la fille que Jean aime vient} produit
l'analyse de la figure~\ref{la_fille_que_}. Cette structure n'est pas
un arbre car elle contient un cycle\footnote{\french{aime} gouverne
  \french{que} car \french{que} est le complément d'objet de
  \french{aime}~;\french{que} gouverne \french{aime} car c'est le
  pronom relatif qui introduit la relative où \french{aime} est le
  verbe.}
et le pronom relatif \french{que} a
deux gouverneurs.

On retrouve ainsi avec notre méthode le problème qui se pose
habituellement en grammaire de dépendances pour gérer les phénomènes
d'extraction. 
Par exemple, si nous reprenons la phrase \french{la fille que Jean
  aime vient}, le mot \french{que} remplit ici deux rôles, il est
l'objet anaphorique de \french{aime} et subordonne la relative à
l'antécédent.  Ce double rôle suppose naturellement deux relations de
dépendances distinctes qui contredisent le principe de représentation
en arbre, alors que l'analyse présentée figure~\ref{la_fille_que_}
rend bien compte de ce double emploi. D'autres approches d'analyses en
dépendances utilise également des structures qui ne sont pas des
arbres~: S.~Kahane~\cite{kahane00} propose une analyse dans laquelle
un pronom relatif a deux gouverneurs~; R.~Hudson~\cite{hudson90}
utilise également souvent des structures dans lesquelles un mot peut
avoir plusieurs gouverneurs.

\subsection{Projectivité}
Une autre question récurrente à propos des structures de dépendances à
considérer pour la description de la langue est celle de la
projectivité. En effet une structure projective induit que les
relations de dépendances restent à un niveau local, ce qui permet une
analyse simple et efficace.

Cette notion initialement définie pour les arbres peut se transposer
sur les graphes~: une structure de dépendances est dite projective si
pour tout mot donné, l'ensemble des n\oe uds atteignables depuis ce
mot dans la structure de dépendances (qu'on appellera {\bf emprise du
  mot}) correspond à un segment continu de l'énoncé.  Par exemple la
structure de la figure~\ref{la_fille_que_} est projective alors que
celle de la figure~\ref{enconnait} ne l'est pas~: dans cette analyse,
l'emprise du mot \french{couleur} est formé de deux segments
\french{en} et \french{la couleur} séparés par le mot
\french{connaît}.

\subsection{Classes de structures de dépendances}
R.~Debusmann and M.~Kuhlmann proposent des critères qui permettent de
classer plus finement les analyses non-projectives. Il obtiennent
ainsi une hiérarchisation en classes du pouvoir expressif que
permettent les différentes structures de
dépendances~\cite{debusmann08}. La notion de {\bf degré de
discontinuité} (block-degree) associe à une structure un entier qui
est le nombre maximum de segments continus disjoints dans l'emprise
d'un mot (un degré de discontinuité de $1$ correspond exactement à la
projectivité).  Pour les structures dont le degré de discontinuité est
au moins $2$, ils distinguent celles qui sont {\bf bien imbriquées}
(well-nestedness) c'est-à-dire celle qui sont telles que les emprises
des deux mots ne se croisent pas (soit elles sont disjointes, soit
l'une est entièrement incluse entre deux segments de l'autre). Sur le
Prague Dependency Treebank, les auteurs montrent que 99,5\% des
analyses sont bien imbriquées et de degré de discontinuité au plus
deux (ce qui est équivalent à être dans la classe de langages des
TAG).

L'application de notre méthode à cette grammaire du français sur la
TSNLP ne produit pas d'analyse mal imbriquées. On obtient dans la
plupart des cas des structures de dépendances projectives. Les
exemples pour lesquels le degré de discontinuité est de $2$ sont dûs
principalement au placement de l'auxiliaire dans le noyau verbal. Les
mots \french{en} ou \french{y} ainsi que l'inversion sujet/verbe lors
de l'emploi de pronoms interrogatifs sont d'autres sources de
discontinuité, mais nous n'avons pas trouvé d'exemple qui aille au-delà
d'un degré de discontinuité de $2$.

\section{Conclusion}
Dans cet article, nous avons proposé une méthode pour construire une
analyse en dépendances d'un énoncé à partir de son analyse en
constituants dans les IG. Cette méthode, basée sur la saturation des
polarités, a mis en évidence deux types de dépendances~: les
dépendances linéaires qui représentent les relations tête-complément ou
tête-spécifieur et les dépendances non-linéaires qui représentent
les relations modifieur-modifié.

Les structures de dépendances obtenues par cette méthode sont des
graphes orientés, elles sont plus riches que les structures obtenues
habituellement par des grammaires de dépendances. Cette représentation
permet de gérer simplement les phénomènes linguistiques posant
habituellement des difficultés dans les grammaires de dépendances.

Pour la suite du travail, nous souhaitons étudier dans quelle mesure
il est possible de transposer les méthodes d'analyses d'un formalisme
à l'autre. Par exemple, Nous avons remarqué que très peu d'analyses
sont non-projectives~; on pourrait donc isoler les cas non-projectifs et
adapter un algorithme d'analyse des grammaires de dépendances qui
servirait de guide à l'analyse dans les IG. Il serait également
intéressant d'étudier l'application de nos méthodes d'analyse
spécifiques aux IG à l'analyse pour des grammaires de dépendances
lexicalisées.  

\bibliographystyle{taln2002} 
\bibliography{taln09}

\end{document}